## TITLE

Biomechanics do not predict perceptual and cognitive consequences of force augmentation

Maria Molina-Sanchez[1*], Francesco Missiroli[2*], Viktorija Pavalkyte[1], Mabel Ziman[1], Julien Russ[1], Maggie Szymanska[1], Noham Wolpe[3], Rani Moran[4], Lorenzo Masia[2#], Tamar R. Makin[1#]

*# Equal contribution

[1] MRC Cognition and Brain Sciences Unit, University of Cambridge, Cambridge, UK; [2] Technical University of Munich, Munich, Germany; [3] Tel Aviv University, Tel Aviv, Israel; [4] University College London, London, UK.

## ABSTRACT

Wearable force-augmentation technologies, including exoskeletons and soft exosuits, are being developed for occupational and rehabilitation settings to enhance performance by reducing the muscular demands of movements. Yet enhanced performance also requires users to perceive the environment accurately, adapt to changing conditions, and decide whether actions are worth their physical cost. We used an elbow-assisting exosuit to investigate the physiological, perceptual and cognitive consequences of force augmentation during lifting. We assessed weight perception, adaptation in a volatile environment, and effort-based decision-making with and without assistance. While assistance reduced muscle activity, its behavioural outcomes did not simply reflect this biomechanical change. Assistance reduced perceived weight, but this effect was strongly shaped by participants' initial experience. Assistance also selectively reduced adaptation to changing environmental contingencies. Finally, reducing the muscular demands of lifting did not translate straightforwardly into how demanding the action felt or whether participants chose to perform it: differences in effort-based decisions were shaped by initial experience rather than simply tracking reductions in muscle activity or perceived fatigue. Together, we show that the perceptual and cognitive consequences of force augmentation cannot be inferred from its biomechanical benefits alone. Effective augmentation therefore requires perceptual and cognitive outcomes to be evaluated alongside physical performance.

## INTRODUCTION

Advances in wearable robotics have enabled the development of force-augmentation devices that reduce the muscular demands of movement. Among these, upper-limb soft exoskeletons (exosuits) provide lifting assistance, with potential applications in occupational and rehabilitation settings (*1*). While their biomechanical benefits are well established (*2*–*5*), successful augmentation requires more than making an action physically easier: users must continue to accurately perceive the properties of objects they interact with, adapt if the environment changes, and decide when an action is worth its physical cost.

People will encounter augmentation after a lifetime of interacting with objects and performing actions without such assistance. Through these interactions, we develop expectations about the physical world and the perceptual consequences of our actions. Force augmentation alters the muscular demands of these familiar interactions without changing the environment itself. An object does not become physically lighter when lifted with an exosuit, yet substantially less muscular activity may be required to lift it. This prompts a key question: how do perceptual and cognitive processes shaped by a lifetime of experience respond when the physical demands of our actions change?

The size-weight illusion illustrates the powerful influence of these expectations: when two objects have identical mass but differ in size, the smaller object is typically perceived as heavier (*6*). Although the mechanisms underlying the size-weight illusion remain debated, mismatch-based accounts propose that the illusion arises from a discrepancy between expected weight and the sensory consequences of lifting an object (*7*). Importantly, size-based expectations continue to bias perceived weight despite repeated experience, even as the forces used to lift the objects adapt to their actual mass (*8*, *9*), and despite explicit knowledge that the objects weigh the same (*10*). So, what happens to weight perception when lifting is assisted? Perceived heaviness has been proposed to reflect not only an object's physical properties, but their relation to the muscular forces required to lift it (*11*). Force augmentation alters this relationship by reducing the muscular demands of lifting while leaving the object itself unchanged. It may therefore change how heavy an object feels and, by changing the sensory information generated during lifting, potentially alter how strongly size-based expectations influence perceived weight (*12*).

The persistence of these expectations highlights a broader challenge for augmentation: how flexibly can we revise what we have learned once the conditions of our interactions with the environment change? Through a lifetime of unassisted action, we learn relationships between our actions, the effort they require, and their consequences. Assistance changes these familiar relationships by allowing the same actions to be performed with less muscular effort. Adapting to augmentation therefore requires the ability to revise expectations formed under unassisted conditions. But does augmentation also affect how flexibly we update what we have learned when relationships in the environment themselves change?

Finally, force augmentation may also change how physically demanding an action feels. However, objective physical demands and their subjective experience are not necessarily equivalent (*13*). For example, people may be willing to undertake greater amounts of physical work without a corresponding change in perceived effort (*14*), while some clinical populations can experience greater perceived effort than healthy individuals under comparable physical demands (*15*, *16*). Likewise, decisions about whether to engage in a physically demanding action depend not only on its objective muscular requirements, but on how its costs are represented relative to its potential benefits (*17–19*). This raises the question of whether reducing muscular demands through force augmentation also reduces perceived fatigue and increases people's willingness to undertake physically demanding actions, or whether these subjective and cognitive consequences remain partly dissociable from its biomechanical effects.

Here, we asked how reducing the muscular demands of lifting through an elbow-assisting exosuit influences perception, learning and effort-based decision-making. Specifically, we examined whether assistance altered perceived weight and the influence of prior expectations on weight judgements; whether it affected the flexibility with which learned relationships were updated when environmental contingencies changed; and whether reductions in muscular demands translated into lower perceived fatigue and changes in effort-reward decision-making. Thirty-two participants completed three tasks with and without exosuit assistance in two separate sessions, with the order of assisted and unassisted sessions counterbalanced across participants. This allowed us to examine both the effects of current assistance and the influence of initial experience on these effects. Surface electromyography (EMG) was recorded to quantify muscle activity during the lifting tasks.

## RESULTS

### Prior experience shapes the effect of exosuit assistance on weight perception

We first examined whether force augmentation distorts weight perception using the size-weight illusion, a well-established phenomenon in which the smaller of two objects of identical mass but different sizes is typically judged as heavier than the larger one because larger objects are expected to weigh more (*6*, *20*). We therefore implemented a bimanual lifting task, adapted from classic size-weight illusion paradigms (*8*, *9*, *21*); see Figure 1A and Movie S1. Participants repeatedly lifted three boxes made of identical material and colour, presented in a pseudo-randomised order: a small box (15 × 15 × 15 cm; 2.850 kg), a large box (25 × 25 × 25 cm; 2.850 kg), and a heavier reference box (25 × 25 × 25 cm; 3.860 kg). Surface EMG was recorded continuously from arm and lower-back muscles. After each lift, participants estimated the box's perceived weight using absolute magnitude estimation, converted to kilograms for analysis.

To investigate the effects of exosuit assistance on muscle activity, we used linear mixed-effects models with box mass (large vs. reference; both of identical size), assistance condition and trial number as predictors of EMG activity. Session order (i.e., whether participants completed the assisted or unassisted condition first) was included to examine whether the effects of assistance were influenced by prior experience with the assisted or unassisted condition, while maximum voluntary force (MVF) was added to account for individual differences in strength. Full model and variable specifications for this and all subsequent analyses are provided in the Statistical Analyses section of the Methods and in Supplementary Tables S2 and S3.

First, we confirmed that biceps EMG activity scaled with box mass (reference − large = 0.024, $F(1, 1354) = 10.67$, $p = .001$) and was reduced by assistance (unassisted − assisted = 0.053, $F(1, 1354) = 90.30$, $p < .001$; Figure 1B), corresponding to an average reduction of approximately 31% relative to the unassisted condition. Importantly, the reduction in muscle activity with assistance did not seem to depend on whether participants experienced the assisted or unassisted condition first (assistance × session order interaction: $F(1, 1354) = 0.27$, $p = .602$). See Figure S1A for the relationship between EMG activity and participants' strength and Figure S1B for EMG analyses of the triceps brachii, deltoid, and lumbar paraspinal muscles.

We next examined whether assistance was associated with changes in perceived weight (Figure 1C), using the same analysis parameters. First, we confirmed that participants were sensitive to true differences in box mass, with higher weight estimates for the heavier reference box than the lighter large box (reference − large = 2.07 kg, $F(1, 1591) = 1622.94$, $p < .001$). This box mass effect did not differ between assisted and unassisted conditions (box mass × assistance interaction: $F(1, 1591) = 2.07$, $p = .15$), suggesting that assistance did not alter participants' discrimination of true differences in box mass. We found a significant main effect of assistance ($F(1, 1591) = 24.71$, $p < .001$), with lower perceived weight in the assisted compared to the unassisted condition. There was no main effect of session order ($F(1, 25) = 0.28$, $p = .601$), indicating that overall weight estimates did not differ according to which condition participants experienced first. However, the assistance condition interacted with session order, indicating that the effect of assistance depended on whether participants experienced the assisted or unassisted session first ($F(1, 1591) = 121.98$, $p < .001$; Figure 1D). See Figure S1C for changes in weight estimates over time.

To determine whether muscle activity contributed to weight perception beyond the effects of session order, we repeated this analysis including EMG activity as a covariate. EMG was not significantly associated with perceived weight ($F(1, 1193) = 2.87$, $p = .091$). Although the main effect of assistance was no longer significant after including EMG ($F(1, 1188) = 2.78$, $p = .096$), the strong assistance × session-order interaction remained highly significant ($F(1, 1177) = 90.81$, $p < .001$). Thus, the influence of prior assisted or unassisted experience on the perceptual effects of assistance could not be explained by differences in muscle activity.

Next, we assessed whether assistance altered size-related biases in weight perception by comparing the small and large boxes of identical mass (i.e., the size-weight illusion). Overall, participants judged the smaller box as heavier than the larger box (small − large = 1.05 kg, $F(1, 1592) = 729.52$, $p < .001$), demonstrating a robust size-weight illusion (Figure 1C). Size-based expectations were also reflected in anticipatory muscle activity, with higher lift-phase EMG activity for the large than the small box (large − small = 0.008, $F(1, 1414) = 5.41$, $p = .020$; see Figure S2B for further details). The size-weight illusion did not significantly differ between assisted and unassisted conditions (box size × assistance interaction: $F(1, 1592) = 0.28$, $p = .59$), indicating that assistance did not alter the influence of size-based expectations on perceived weight (Figure 1E). Instead, we found a significant main effect of assistance on overall perceived weight across both boxes ($F(1, 1592) = 19.97$, $p < .001$), with lower perceived weight in the assisted compared to the unassisted condition. Again, this effect interacted with session order ($F(1, 1592) = 198.95$, $p < .001$; Figure S2A), such that the effect of assistance on perceived weight depended on which condition participants experienced first.

Together, these results show that while exosuit assistance reliably reduced muscle activity, it preserved both sensitivity to true differences in box mass and the influence of size-based expectations on perceived weight. Assistance reduced overall weight estimates, but this effect depended strongly on prior experience. This influence of prior experience persisted after accounting for differences in muscle activity, indicating that it could not be explained by the altered muscular demands of lifting alone.

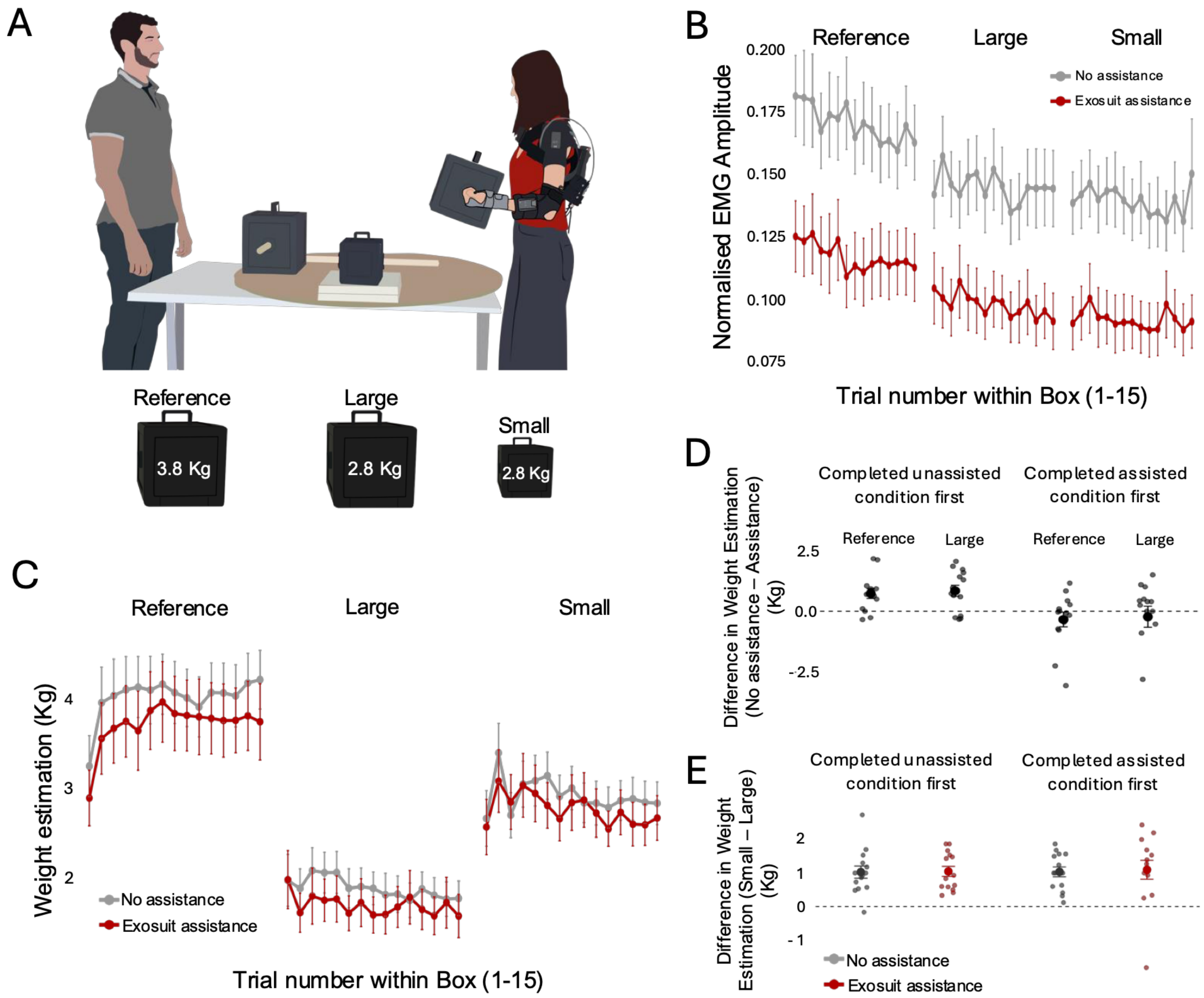

***Figure 1. Effects of exosuit assistance on weight perception.***

*(**A**) Experimental setup and stimuli. Participants (n=31) repeatedly lifted three boxes with and without assistance: reference (25 × 25 × 25 cm; 3.860 kg), large (25 × 25 × 25 cm; 2.850 kg), and small (15 × 15 × 15 cm; 2.850 kg). (**B**) Mean normalised biceps EMG activity across trials for each box in the assisted (red) and unassisted (grey) conditions. Assistance reduced EMG activity across all boxes. (**C**) Mean perceived-weight estimates across trials. Participants distinguished the heavier reference box from the lighter large box and perceived the small box as heavier than the large box, consistent with the size-weight illusion. Overall, weight estimates were lower in the assisted condition. (**D**) Difference in perceived weight between the unassisted and assisted conditions for the reference and large boxes, separated by session order. Positive values indicate higher perceived weight without assistance. The effect of assistance differed according to which condition participants experienced first. (**E**) Difference in perceived weight between the small and large boxes (size-weight illusion), separated by session order. Illusion magnitude was comparable between assisted and unassisted conditions. Larger dots and error bars represent group mean ± SEM. Small dots in panels D and E represent individual participants.*

**Assistance alters behavioural flexibility without changing fitted learning rates**

We next examined whether exosuit assistance influenced adaptive learning about object weight under uncertainty. To assess this, participants completed a probabilistic cue-outcome learning task involving multiple contingency reversals. On each trial, a box labelled with one of two tags (A or B) was presented. Participants were informed that one tag predicted an 80% probability that the box would be heavy, whereas the other predicted an 80% probability that it would be light, although these associations could reverse throughout the task. Before lifting the box, participants indicated whether they expected the box to be heavy or light (see Figure S3 and Movie S2 for task schematics).

To characterise behavioural adaptation across contingency blocks, we modelled trial-level accuracy using a binomial mixed-effects model (Figure 2A). Adaptation was quantified as the change in accuracy across trials within each block ("trials since reversal"). Accuracy increased as a function of trials since reversal ($\chi^2(1) = 397.15$, $p < .001$), indicating that participants adapted to the new contingencies over time. Although there was no main effect of assistance on accuracy ($\chi^2(1) = 0.24$, $p = .626$), adaptation dynamics differed between assisted and unassisted conditions (assistance × trials-since-reversal: $\chi^2(1) = 6.88$, $p = .009$). The effect of assistance on adaptation varied between contingencies (assistance × trials-since-reversal × contingency: $\chi^2(1) = 4.78$, $p = .029$). There was no significant main effect of session order ($\chi^2(1) = 0.43$, $p = .513$), nor an assistance × session order interaction ($\chi^2(1) = 0.02$, $p = .889$), providing no evidence that the influence of assistance on learning in a volatile environment depended on which condition participants experienced first, unlike its effects on weight perception. Repeating the analysis after excluding the initial learning block (Block 1) yielded similar results, indicating that the effects persisted when only actual reversals were considered.

Follow-up slope comparisons showed that accuracy increased more steeply in the unassisted than in the assisted condition for the A-light contingency (unassisted − assisted = 0.126, $z = 3.59$, $p < .001$), whereas adaptation slopes did not significantly differ between assistance conditions for the B-light contingency (unassisted − assisted = 0.012, $z = 0.30$, $p = .766$).

To examine whether the effect of assistance on adaptation differed across individual contingency blocks, we conducted a follow-up block-level analysis. The assistance × trials-since-reversal × contingency-block interaction did not reach statistical significance ($\chi^2(4) = 7.83$, $p = .098$). Exploratory follow-up comparisons showed significantly steeper adaptation in the unassisted than assisted condition in block 3 (unassisted − assisted = 0.156, $z = 2.50$, $p$Holm = .049) and block 5 (unassisted − assisted = 0.184, $z = 2.83$, $p$Holm = .023), but not at blocks 1, 2, or 4 (all $p$Holm ≥ .685, Figure 2B). This pattern suggests that exosuit assistance may have selectively reduced behavioural flexibility when the A-light contingency was reintroduced.

To investigate the learning processes underlying these behavioural differences, we fitted four reinforcement-learning models to participants' expectancy responses (see the Computational Modelling section in Methods for model specifications, fitting and comparison procedures). The opponent Rescorla-Wagner (RW) model provided the best fit in 55 of 64 participant-sessions (27 unassisted and 28 assisted). We therefore compared the model-derived learning-rate parameter between assisted and unassisted conditions. Learning rates did not differ significantly between conditions (paired Wilcoxon signed-rank test, $W = 260$, $p = .940$; Figure 2C). Simulations using the fitted opponent RW model reproduced the direction, but not the magnitude, of the assistance-related reduction in adaptation for the A-light contingency (see Supplementary Results).

Together, these findings indicate that exosuit assistance altered behavioural adaptation across repeated contingency reversals without a detectable change in model-derived learning rates. Although participants learned the cue-outcome associations successfully under both conditions, assistance selectively reduced behavioural flexibility when the A-light contingency was reintroduced after reversal experience. This suggests that the effects of augmentation on adaptive learning depended on the contingency being learned, rather than on which assistance condition participants experienced first.

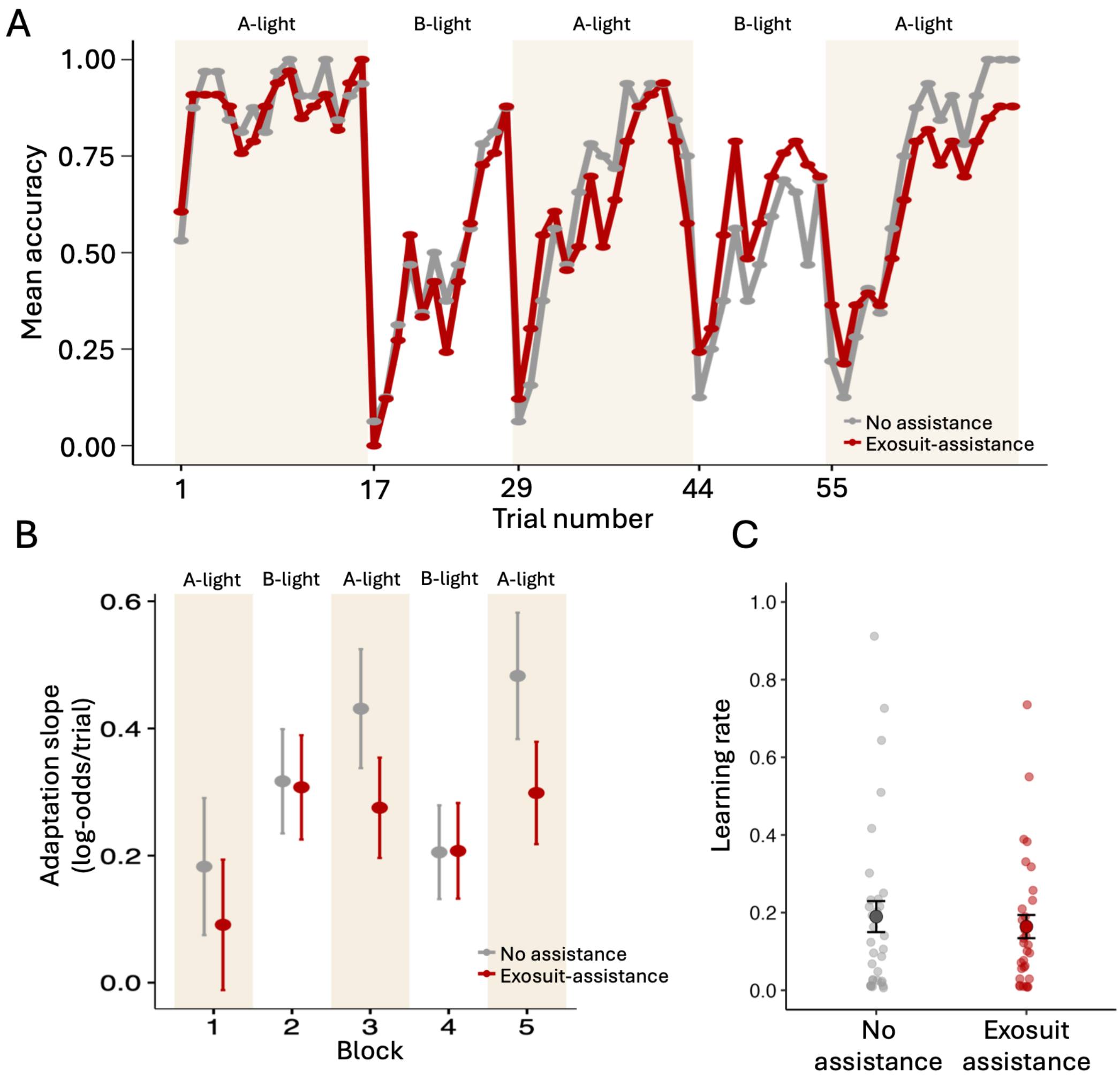

***Figure 2. Effects of exosuit assistance on learning and adaptation.***

*(**A**) Mean accuracy across trials for participants (n = 32) in the unassisted (grey) and assisted (red) conditions. Shaded regions indicate A-light contingencies; unshaded regions indicate B-light contingencies. (**B**) Model-estimated adaptation slopes for each contingency block, representing the change in log-odds of a correct response per trial during the first 11 trials of each block. Dots indicate model-estimated slopes and error bars denote 95% confidence intervals. Exploratory comparisons showed lower adaptation slopes under exosuit assistance at blocks 3 and 5, when the A-light contingency was reintroduced. (**C**) Learning rates estimated using the opponent Rescorla-Wagner model. Large dots indicate group means and error bars denote ± SEM; small dots represent individual participants. Learning rates did not significantly differ between the unassisted and assisted conditions.*

**Assistance reduces perceived fatigue, but fatigue does not directly track muscle activity**

To assess how exosuit assistance influenced fatigue and effort-related decision-making, participants completed an effort-reward decision-making task adapted from previous studies (*22–24*). On each trial, participants chose between a low-effort, low-reward option (light box: 1.33 kg; reward: £1) and a high-effort, high-reward option (heavy box: individually calibrated to 35% of each participant's MVF; reward varied from £3 to £7), before lifting the selected box (Figure 3A and Movie S3). The task comprised 60 trials divided into four equal blocks. EMG activity was recorded during each lift. To track how perceived fatigue evolved over time, participants provided verbal fatigue ratings after each trial on a scale from 0 ("completely rested") to 100 ("worst imaginable fatigue").

We first confirmed the effect of exosuit assistance on biceps muscle activity during lifting in the subset of participants with available EMG data (n = 21). EMG activity was analysed using a linear mixed-effects model with assistance condition, trial number and session order as predictors. The load lifted was included as a covariate to account for differences in the weights participants chose across trials and conditions. EMG activity was lower in the exosuit-assisted condition (unassisted − assisted = 0.032, $F(1, 2373.0) = 6.04$, $p = .014$; Figure 3B) and increased with the load lifted ($\beta = 0.026$ per kg, $F(1, 2379.9) = 3238.32$, $p < .001$). EMG also increased over the course of the task ($F(1, 2373.0) = 28.77$, $p < .001$). However, there was no significant assistance × trial number interaction ($F(1, 2373.0) = 2.38$, $p = .123$), indicating that the change in EMG activity over time did not differ significantly between conditions. The effect of assistance was also modulated by session order ($F(1, 2373.1) = 6.25$, $p = .012$). EMG activity was lower with assistance in both groups, but this reduction was larger when participants experienced the assisted condition first (both *p* values $< .001$).

We next examined perceived fatigue during the same task using the same predictors, with perceived fatigue immediately preceding the effort-reward task additionally included as a covariate to account for differences in fatigue at task onset. Perceived fatigue was lower in the assisted condition ($F(1, 3248.6) = 117.77$, $p < .001$) and increased over the course of the task ($F(1, 3233.2) = 1350.90$, $p < .001$). However, the rate of increase differed between conditions (assistance × trial interaction: $F(1, 3233.1) = 22.26$, $p < .001$), such that the difference in perceived fatigue between conditions decreased over the course of the task (unassisted − assisted: 5.05 fatigue points at trial 1 vs. 1.55 fatigue points at trial 60; Figure 3C). The effect of assistance also differed according to session order ($F(1, 3253.3) = 68.99$, $p < .001$), with a larger reduction in perceived fatigue among participants who completed the unassisted session first, although fatigue was significantly lower with assistance in both groups (both $p$ values < .001).

Finally, we examined whether perceived fatigue directly reflected the muscle activity required during each lift by testing whether trial-by-trial EMG activity predicted the change in perceived fatigue from one trial to the next. EMG activity during the current lift was not significantly associated with changes in perceived fatigue ($F(1, 82.28) = 0.36$, $p = .553$), and this relationship did not differ significantly between assisted and unassisted conditions (assistance × EMG interaction: $F(1, 1360.17) = 1.37$, $p = .242$). In contrast, heavier loads were associated with greater trial-by-trial increases in perceived fatigue ($\beta = 0.189$ per kg, $F(1, 366.20) = 193.24$, $p < .001$). Thus, changes in perceived fatigue were sensitive to the physical load being lifted but were not directly predicted by the magnitude of muscle activity during that lift.

Overall, these findings show that exosuit assistance reduced perceived fatigue throughout the task despite a faster rate of fatigue accumulation over time. Trial-by-trial changes in perceived fatigue were strongly related to the physical load being lifted but not significantly associated with muscle activity during those lifts.

**Assistance alters effort-based choice while preserving reward sensitivity**

We next asked whether the reductions in muscle activity and perceived fatigue observed with exosuit assistance were reflected in effort-based decisions. Specifically, we examined whether participants were more willing to choose the high-effort option under assistance and whether any behavioural differences emerged over the course of the task as perceived fatigue accumulated.

To examine effort-based behaviour, we modelled trial-by-trial choices (high- versus low-effort option) using a binomial mixed-effects model with assistance condition, trial number and session order as predictors. Reward magnitude was included as a covariate. Higher rewards strongly increased the likelihood of choosing the high-effort option ($\chi^2(4) = 725.72$, $p < .001$), whereas participants became progressively less likely to choose the high-effort option as the task progressed ($\beta = -0.013$, $\chi^2(1) = 16.17$, $p < .001$). The effect of assistance on effort-based choices changed over time (assistance × trial: $\chi^2(1) = 12.97$, $p < .001$), and this time-dependent effect depended on session order (assistance × trial × session order: $\chi^2(1) = 6.56$, $p = .010$). Follow-up analyses showed that the assistance × trial interaction was significant when participants experienced the assisted session first ($p < .001$), but not when they experienced the unassisted session first ($p = .399$; Figure 3D). Thus, despite lower muscle activity and perceived fatigue under assistance from the beginning of the task, its effect on effort-based choices emerged gradually in participants who experienced assistance first. No significant difference in how choices evolved over time was observed in those who experienced the unassisted condition first.

We next examined whether changes in effort-based choices over time were associated with accumulated muscular demands and perceived fatigue. In the subset of participants with available EMG data ($n = 21$), we tested whether cumulative EMG activity across all preceding trials and perceived fatigue on the previous trial predicted participants' subsequent choice. Greater cumulative EMG activity ($\beta = -0.065$, $\chi^2(1) = 14.18$, $p < .001$) and higher previous-trial fatigue ($\beta = -0.034$, $\chi^2(1) = 16.75$, $p < .001$) were both associated with a lower likelihood of choosing the high-effort option on the subsequent trial. However, the effect of assistance on how choices changed over time remained significant after accounting for both cumulative EMG activity and perceived fatigue (assistance × trial number: $\chi^2(1) = 6.88$, $p = .009$). Thus, although accumulated muscular demands and perceived fatigue were associated with greater effort avoidance, they did not fully explain the effect of assistance on effort-based decision-making over time.

Finally, we asked whether reducing the muscular demands of the task altered how participants balanced reward against effort during decision-making. If exosuit assistance reduced the subjective effort cost of the high-effort option, the influence of reward magnitude on choice might also change. To test this, we fitted a generalised linear mixed-effects model including reward magnitude, assistance condition, trial number and session order as predictors of choice. As expected, higher rewards increased the likelihood of choosing the high-effort option ($\beta = 1.51$, $\chi^2(1) = 719.27$, $p < .001$). However, reward sensitivity did not differ between assisted and unassisted conditions (reward × assistance: $\beta = -0.060$, $\chi^2(1) = 1.67$, $p = .196$, BIC-approximated $BF_{01} = 26.08$; Figure 3E), nor did it change over the course of the task (reward × trial: $\beta = -0.0002$, $\chi^2(1) < 0.01$, $p = .946$). There was also no evidence that the effect of assistance on reward sensitivity changed over time (reward × assistance × trial: $\beta = 0.002$, $\chi^2(1) = 0.56$, $p = .456$). These findings indicate that exosuit assistance did not alter how strongly reward magnitude influenced effort-based choices.

Together, these results indicate that although accumulated muscular demands and fatigue were related to effort-based decisions, they did not fully account for the time-dependent influence of exosuit assistance on choice, which appeared to depend on participants' initial experience with the device. In contrast, sensitivity to reward was comparable across conditions, indicating that the influence of reward on choice was preserved under assistance.

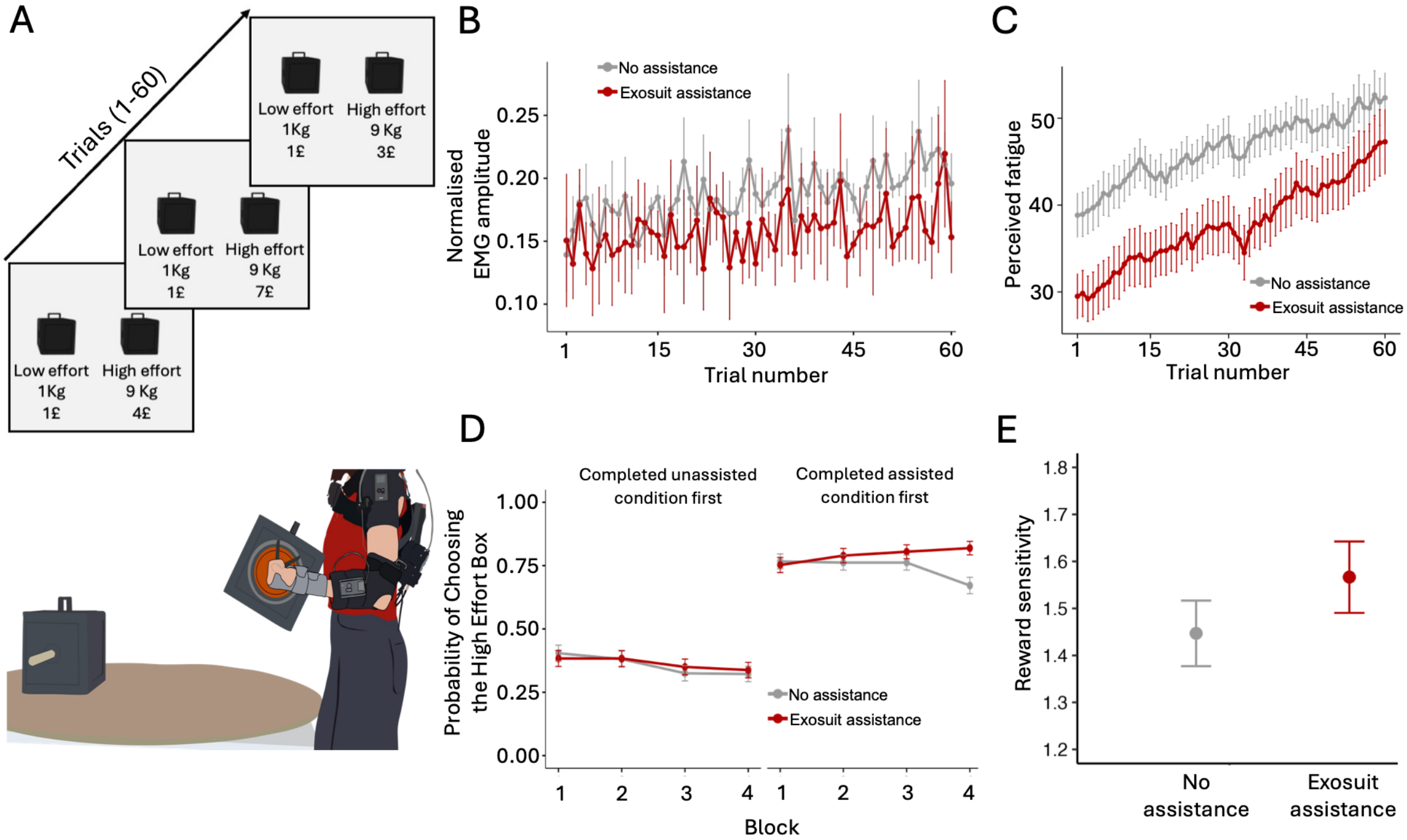

***Figure 3. Effects of exosuit assistance on perceived fatigue and effort-based decision-making.***

*(**A**) Task design and experimental set-up. On each trial, participants (n = 30) chose between a low-effort/low-reward option and a high-effort/high-reward option. (**B**) Trial-by-trial normalised EMG amplitude during the task, adjusted for the weight lifted. Exosuit assistance (red) reduced overall muscle activity throughout the task, compared to the unassisted condition (grey). Points and error bars represent mean ± SEM (n = 21). (**C**) Trial-by-trial perceived fatigue during the task. Fatigue increased progressively over time in both conditions but remained significantly lower under exosuit assistance. However, fatigue accumulated more rapidly in the assisted condition. Points and error bars represent mean ± SEM. (**D**) Proportion of high-effort choices across task blocks, split by session order. High-effort choices declined without assistance but remained stable with assistance among participants who completed the assisted condition first. Points and error bars represent mean ± SEM. (**E**) Reward sensitivity did not differ between assisted and unassisted conditions. Points and error bars represent model-estimated slopes ± SE.*

## DISCUSSION

Force augmentation changes the muscular demands of movement, but our findings show that its perceptual and cognitive consequences do not simply follow from this biomechanical change. Across perception, learning and decision-making, the effects of assistance were selective and, in some cases, strongly shaped by the context established through participants' initial experience of the task. This reveals an important challenge for augmentation: changing what an action physically requires does not necessarily tell us how people will experience or respond to that change.

An important feature of our findings was that the effects of augmentation were not determined solely by whether assistance was currently present or absent, but also by the condition in which participants first experienced the task. Because these tasks were new to participants and assistance order was counterbalanced, we could examine whether assistance effects differed depending on whether the initial session was assisted or unassisted. This influence was clearest for weight perception. The effect of assistance on perceived weight depended strongly on which assistance condition participants experienced first, even though the reduction in muscle activity during the weight-estimation task did not. Moreover, this influence of initial experience remained after accounting for muscle activity, indicating that it was not attributable to differences in muscular demands. Thus, the perceptual consequences of assistance depended on previous experience even when its effect on muscular demands did not.

Beyond weight perception, initial experience also influenced subsequent effects of assistance, although not always in the same way: it modulated the magnitude of the reduction in perceived fatigue and muscle activity during effort-based decision-making, while also altering how effort-based choices evolved over time. The perceptual and cognitive effects of augmentation therefore depended partly on the sequence in which assisted and unassisted states were experienced, rather than solely on the current assistance state. One possibility is that the condition in which an action is initially performed establishes a reference that shapes subsequent responses to that action under assisted or unassisted conditions (*25*, *26*). Notably, these effects persisted despite the sessions being separated by at least two weeks, suggesting that initial experience established a relatively persistent context for subsequent performance. This influence of prior experience could have important implications for transitions between assisted and unassisted work, where responses to one assistance state may continue to be shaped by experience with the other (*27*).

Interestingly, the effect of augmentation on learning did not depend on which assistance condition participants experienced first. Instead, assistance selectively reduced behavioural flexibility when participants had to readapt to a previously encountered contingency. This may reflect an important difference from the perceptual and effort-based effects: in this volatile environment, learning required continuous updating in response to changing contingencies, potentially making immediate trial-by-trial experience more relevant than the context established in the previous session. However, this reduction in flexibility emerged only when one of the previously learned contingencies was reinstated, rather than consistently across reversals. It therefore appears to reflect a selective effect of assistance on readaptation, although why it emerged for this particular contingency remains unclear.

A second important feature of our findings was that the biomechanical, subjective and decision-making effects of augmentation did not map straightforwardly onto one another. Assistance substantially reduced both muscle activity and perceived fatigue. However, these overall reductions did not imply that perceived fatigue closely tracked muscular demands: muscle activity was not significantly associated with trial-to-trial changes in fatigue after accounting for the load lifted. Thus, participants' fatigue ratings seemed to be updated according to the physical load they chose to lift, rather than closely tracking the muscular demands required to lift it. Together with previous evidence of dissociations between physiological and subjective responses to exoskeleton assistance (*28*, *29*), this suggests that subjective responses to assistance cannot be inferred from physiological changes alone. Perceived fatigue may also be informed by motivational, affective or contextual factors (*13*, *30*). For example, expectations that assistance will make an action easier may contribute to how fatiguing that action feels, beyond the reduction in muscular demands itself (*31*).

Perceived fatigue and muscular demands were, in turn, related to effort-based decisions, but did not fully account for how assistance influenced choices over time. Greater accumulated muscle activity and greater perceived fatigue were associated with a lower likelihood of choosing the high-effort option, yet the time-dependent effect of assistance on choice remained after accounting for both. Moreover, changes in muscle activity and perceived fatigue did not map directly onto effort-based choices. When participants experienced the unassisted condition first, assistance reduced muscle activity and perceived fatigue without altering how choices evolved over time. When participants experienced assistance first, the difference in perceived fatigue between conditions became progressively smaller, yet their choices increasingly diverged: high-effort choices were maintained under assistance but declined without it. Importantly, reward sensitivity remained stable across assistance conditions, suggesting that this divergence was not driven by a change in how strongly reward influenced choice.

It is possible that maintaining high-effort choices under assistance contributed to the faster accumulation of perceived fatigue, as heavier loads were associated with greater fatigue. Importantly, however, the faster increase in perceived fatigue under assistance was observed while accounting for the load lifted on each trial, indicating that differences in choice alone were unlikely to fully account for this effect. Previous work on perceived effort suggests that perceiving an action as more effortful does not necessarily lead to greater effort avoidance, with choices influenced by factors beyond perceived effort alone (*14*, *32*, *33*). Our findings suggest that a similar dissociation may occur for perceived fatigue, which did not straightforwardly map onto effort-based choices. For augmentation technologies, this means that reducing the physical demands of an action cannot be assumed to produce proportional changes in how demanding that action feels or in whether users choose to perform it.

Overall, our findings show that the perceptual and cognitive consequences of force augmentation cannot be predicted from reductions in muscular demands alone and that these effects are often shaped by the context established through prior experience. An important limitation, however, is that participants experienced exosuit assistance during a single experimental session, and we therefore cannot determine whether these effects persist or change with prolonged exposure. Previous work has shown that the sensorimotor and perceptual consequences of wearable augmentation can change with training (*34*–*36*). Future studies should examine how our perception of the world and interactions with it evolve with prolonged experience of force augmentation and whether the influence of initial experience persists or diminishes over time. Our findings therefore highlight the need to understand augmentation not only in terms of how it changes the physical demands of action, but also how those changes are experienced and translated into behaviour.

# MATERIALS AND METHODS

## Participants

We recruited thirty-two healthy adults (19 female; mean age 24.56 ± 5.31 years; mean height 1.69 ± 0.07 m; mean weight 64.20 ± 12.15 kg) via online advertisements through the MRC Cognition and Brain Sciences Unit and the University of Cambridge SONA participant pools.

Participants completed a screening questionnaire to exclude cardiovascular or musculoskeletal conditions that could affect lifting performance. Individuals were also excluded if they trained in weightlifting regularly, had jobs involving routine lifting, or were pregnant.

All participants provided written informed consent, and the study was approved by the Cambridge Psychology Research Ethics Committee (PRE.2023.066).

An additional five individuals were recruited but did not complete the study: two owing to excessively long intervals between sessions, one who was fasting during a session, and two who experienced vasovagal syncope or presyncope during testing (see Supplementary Table S1 for per-task sample sizes and additional exclusions).

## Bimanual upper-limb soft exoskeleton (exosuit)

We employed a bimanual, soft, and fully wearable robotic system designed to assist elbow flexion in both arms while maintaining natural, unconstrained upper-limb motion. It consists of two mirrored units, one for each arm, derived from the modular soft-exosuit architecture described by  Lotti et al. (*37*). Each side operates independently yet symmetrically, allowing coordinated bilateral movements and providing balanced mechanical support during daily tasks.

### Mechanical and Structural Design

The exosuit consists of: (i) a custom textile harness that wraps around the upper arm and forearm, with anchor points strategically positioned near the elbow to transmit assistive torque, (ii) a back-mounted actuation and control module that serves both arms and houses the motors, controllers, batteries, and communication electronics; and (iii) a pair of Kevlar tendons routed through Bowden sheaths (Shimano SLR, 5 mm) that connect the actuators to the anchor points. These artificial tendons act as “cables” transmitting pulling forces that generate elbow flexion while allowing free extension when slackened. The tendon system minimises mechanical resistance and friction, enabling smooth transmission of forces while keeping the system compliant and comfortable. The 3D-printed distal anchor is integrated into the forearm orthosis, while the proximal anchor at the shoulder provides leverage for effective torque generation. The combined weight of the actuation and control module is approximately 2 kg, with battery-powered operation supporting about 8 hours of active use.

Actuation and Sensing

Each side of the exosuit is driven by a flat brushless DC motor (T-Motor AK60-6, 24 V, 6:1 planetary gearbox) that reels the cable tendons to generate controlled pulling forces. To monitor the user’s motion and interaction with the device, each arm is equipped with (i) two inertial measurement units (IMUs) (Bosch BNO055) mounted on the upper arm and forearm to capture 3D orientation and joint kinematics, and (ii) a force sensor (ZNLBM-1, 20 kg max load) placed in the tendon line to measure interaction forces between the exosuit and the limb. All sensors communicate wirelessly through Bluetooth Low Energy (BLE) to the embedded control electronics.

Control and Operation

The control architecture operates in real time and is based on the *Dynamic Arm Module* framework. It uses a biomechanical model of each arm, customised to the wearer’s anthropometry, to estimate the torque needed to counteract gravity and partially compensate for limb dynamics. The system runs an admittance controller, which continuously compares the measured interaction torque with the reference torque and adjusts the motor velocity to deliver smooth, user-synchronous assistance (*3*).

This adaptive and calibration-free control enables the exosuit to respond naturally to the wearer’s self-initiated movements, supporting voluntary elbow flexion rather than imposing a predefined trajectory. Because each arm’s controller operates independently, assistance can be asymmetric or bilateral, depending on the user’s needs or task demands.

**Experimental design**

We used a within-subject, two-session design. In one session, participants wore the exosuit with powered assistance; in the other, they wore the same exosuit with motors disabled. Session order was randomised and counterbalanced (assisted→unassisted: $n = 16$; unassisted→assisted: $n = 16$). The groups did not significantly differ in age ($W = 129.00$, $p = .985$, $BF_{10} = 0.350$), weight ($t(30) = -0.510$, $p = .615$, $BF_{10} = 0.371$), height ($t(30) = -1.062$, $p = .30$, $BF_{10} = 0.517$), handedness ($\chi^2(1) = 0.00$, $p = 1.000$, $BF_{10} = 0.285$), or proportion of females ($\chi^2(1) = 3.239$, $p = .072$, $BF_{10} = 1.905$). Sessions were separated by at least 15 days (assisted→unassisted: 22.467 ± 7.415 days; unassisted→assisted: 23.938 ± 5.092 days; mean ± SD) to minimise immediate carry-over effects between sessions. There were no between-group differences in inter-session intervals (Kolmogorov–Smirnov test: $D = 0.280$, $p = .520$).

Before starting the experimental protocol, each participant's maximum voluntary force (MVF) was assessed to quantify individual maximal isometric strength. The assessment was performed while participants wore the exosuit to account for the device's intrinsic weight and any mechanical impedance introduced by the suit. Participants were seated comfortably with the trunk upright, the tested arm positioned parallel to the torso, and the elbow flexed at 90°. The forearm remained in a neutral position (midway between pronation and supination), with the wrist aligned to prevent compensatory shoulder motion. A padded cuff was fastened around the participant's wrist and connected via a Kevlar line to a tension load cell, which measured the exerted pulling force during the trial.

Participants performed a maximal voluntary isometric contraction of elbow flexion by pulling against the fixed cable while avoiding shoulder or trunk compensation. Each participant completed three repetitions lasting approximately 3–5 s, with at least 1 min of rest between trials to minimise fatigue. The maximum recorded force across the three trials was taken as the participant's MVF and expressed in newtons.

Each experimental session lasted approximately 3 h and comprised three experimental tasks involving bimanual box lifting in the following fixed order: (1) perceptual weight estimation task, (2) probabilistic cue–outcome learning task, and (3) effort–reward decision-making task. For all tasks, participants stood facing a circular rotating table. An experimenter positioned on the opposite side rotated the table so that, before each lift, the box was centred directly in front of the participant. This helped maintain a consistent foot position, body orientation, and reach distance throughout the experiment. The distance between the participant and the table was adjusted so that, when grasping the box handles, the elbow was slightly flexed (~15° from full extension). Shorter participants stood on a platform, whereas taller participants stood directly on the floor, ensuring a consistent starting posture across individuals. Participants also wore wrist restraints that limited flexion–extension movements, reducing wrist-based cues to perceived weight and focusing effort on the elbow flexors (biceps). Throughout the tasks, surface EMG was recorded from biceps brachii, triceps brachii, deltoid, and lumbar paraspinals.

Before each lift, an auditory cue ("get ready") instructed participants to stand in position, grasp the ends of the rod, and prepare without lifting. A second cue ("lift") signalled them to raise the box within 1 s until the elbow reached ~90° (upper arm and forearm perpendicular to each other). Following a "hold" cue, the box was held at ~90° for 3 s until the "place down" cue prompted participants to return the box to the table. EMG recording began immediately before the "lift" cue and continued until 2 s after the "place down" cue. At the start of each session, after the exosuit was fitted, participants practised this lifting technique using the auditory cues employed throughout the tasks. Practice trials used a visually similar box with a different weight from those used in the experimental blocks.

Perceived fatigue was rated on a 0–100% scale (0 = completely rested; 100 = the worst imaginable fatigue) at the start of each session, after Task 1 and Task 2, and multiple times during Task 3. At the end of the experiment, participants completed two questionnaires assessing embodiment and perceived functionality.

**Stimuli: Boxes**

Three 3D-printed black boxes were used in the study: one small box (15 × 15 × 15 cm) and two large boxes (25 × 25 × 25 cm; Figure 1A). Apart from size, the boxes were identical in material and colour to avoid material-based cues during weight estimation (*38*, *39*). Each box had a wooden rod (57 cm length × 2 cm diameter) passing laterally through its midline and protruding symmetrically on both sides. Box mass was manipulated by attaching U-shaped iron ring weights to the rod that passed through the box; rings were fixed at the rod's midpoint and secured to prevent movement, keeping the centre of mass at the geometric centre. Each box had a magnetically fastened lid, allowing the experimenter to add/remove weights during the experiment, without revealing the load. In the small box, 0.5-kg iron rings were used, resulting in a total mass of 2.850 kg, including the rod. The total mass of the large boxes varied across tasks depending on the number of rings added. For the decision-making task, when additional mass was required, circular iron rings were mounted symmetrically on the rod outside the box.

**EMG recording**

Surface EMG was recorded using 24-mm bipolar electrode pairs (CDE-C, OT Bioelettronica, Torino, Italy) placed directly on the skin over the muscle belly, aligned with the presumed muscle fibre direction. Electrode placement followed the SENIAM recommendations to ensure consistency in inter-electrode spacing, orientation, and anatomical positioning. Signals were acquired with a portable wireless amplifier (Sessantaquatro+, OT Bioelettronica) connected via an eight-channel bipolar adapter (AD8x1SP) using four 3.5-mm four-pole jacks. The amplifier streamed data via Bluetooth to OTBioLab+ software (v1.6.0.0) for real-time visualisation, and recordings were stored for subsequent offline analysis. The reference electrode (CPAT1) was positioned over the cervical spine. The amplifier unit was secured to the participant in a small pouch, and electrode leads (ADx2SP cables) were strain-relieved and taped to the skin using kinesiology tape (Sporttape) to minimise motion artefacts.

EMG activity was collected from four muscles on the right side, each recorded on a separate bipolar channel: channel 1: biceps brachii - long head (elbow flexor); channel 2: triceps brachii - long head (elbow extensor); channel 3: anterior deltoid (shoulder flexor); channel 4: lumbar paraspinals (postural stabilisers). Electrode locations were identified using anatomical landmarks and palpation to locate the muscle belly. Signals were sampled at 2000 Hz, amplified, band-pass filtered (10–500 Hz), and digitised with a 16-bit A/D converter (3-dB bandwidth within the stated passband).

Before starting the experimental tasks, maximum voluntary contractions (MVCs) were collected from each recorded muscle while participants wore the exosuit. MVCs were used for signal normalisation, to account for inter-subject variability and the influence of the device's weight. Offline processing was performed in MATLAB (MathWorks, Natick, MA, USA): EMG signals were filtered with a fourth-order Butterworth filter (15–450 Hz), rectified, and low-pass filtered at 6 Hz (fourth-order Butterworth) to extract the linear envelope. Each envelope was normalised to the participant's MVC. The normalised EMG amplitude was then used as an index of muscle effort to evaluate the effect of exosuit assistance on muscular activation. EMG data from the probabilistic cue–outcome learning task were not included in the analyses because intermittent disconnections during data acquisition resulted in insufficient usable data.

**Task 1: Perceptual weight estimation task**

Participants completed a bimanual perceptual weight estimation task, adapted from classic size–weight illusion paradigms (*8*, *9*, *21*), to assess how exosuit assistance influenced perceived weight (see Movie S1). Participants lifted three boxes: a small box (15 × 15 × 15 cm; 2.850 kg), a large box (25 × 25 × 25 cm; 2.850 kg), and a reference box (25 × 25 × 25 cm; 3.860 kg). During lifts, participants grasped the protruding rod ends with both hands, placing their hands equidistant from the box centre to keep hand spacing constant and minimise haptic size cues in weight judgments (*40*, *41*). To equalise rod height and avoid posture differences, the small box was elevated on three stacked platforms, whereas each large/reference box sat on a single platform. At the start of each session, participants briefly held a 1-kg ring weight with both hands to provide a reference for 1 kg.

Participants completed 45 lifts in total. Lifts were presented in a pseudo-randomised order constrained so that each block of three lifts contained one lift of each box; thus, each box was lifted 15 times. Before each lift, an auditory cue ("get ready") instructed participants to stand in position, grasp the ends of the rod, and prepare without lifting. A second cue ("lift") signalled them to raise the box within 1 s until the elbow reached ~90° (upper arm and forearm perpendicular). Following a "hold" cue, the box was held at ~90° for 3 s until the "place down" cue prompted return to the table. EMG recording began immediately before the "lift" cue and continued until 2 s after the "place down" cue.

After placing the box down, participants rated its perceived weight using absolute magnitude estimation, where larger numbers indicated heavier weights and smaller numbers lighter ones (*42*). Estimates were provided on a self-chosen numerical scale ($n$ = 30 used kilograms and $n$ = 1 used pounds, which was converted to kilograms for analysis). They were free to choose their own unit of measurement and were required to use it consistently across all lifts. Participants were explicitly instructed to report the weight they experienced, rather than what they expected or believed to be correct.

**Task 2: Probabilistic cue-outcome learning task**

Participants completed a probabilistic cue–outcome learning task designed to assess associative learning and behavioural adaptation under contingency reversals (see Movie S2). The task consisted of 70 trials divided into five contingency blocks of 16, 12, 15, 11, and 16 trials, respectively, separated by four unsignalled reversals. On each trial, one of two visual tags ("A" or "B") was attached to the front of a box. Cue presentation was pseudorandomised, with the two cues occurring equally often in even-length blocks and approximately equally often in odd-length blocks. Across the 70 trials, each cue was presented 35 times. Participants were informed that one tag predicted an 80% probability that the upcoming box would be heavy, whereas the other predicted an 80% probability that it would be light, although these cue–outcome associations could reverse throughout the task. Before each lift, participants verbally indicated whether they expected the upcoming box to be "heavy" or "light". They then lifted the box and experienced its actual weight but were not required to verbally report whether the experienced outcome matched their prediction.

The light box weighed 1.850 kg, and the heavy box weighed 2.350 kg. Before starting the task, while exosuit assistance was disabled, participants lifted both boxes to ensure that they could reliably distinguish between the light and heavy weights used during the task. During the task, the experimenter, positioned on the opposite side of the rotating table, opened a magnetic lid on the box to add or remove weights between trials without revealing the load to the participant. A curtain hanging between the participant and the experimenter prevented participants from seeing the weight manipulation procedure. The lifting procedure was identical to that used in Task 1. EMG activity was recorded throughout the task; however, due to technical issues and substantial data loss, these recordings were not analysed further.

The sequence of cue–outcome pairings was identical across participants and sessions. Contingencies alternated between "A-light" blocks (blocks 1, 3 and 5), in which tag A predicted the light box with 80% probability, and "B-light" blocks (blocks 2 and 4), in which tag B predicted the light box with 80% probability. Within each block, probabilistic violations occurred on approximately 20% of trials to maintain uncertainty and require continuous updating of cue–outcome expectations.

Participants were instructed to learn the probabilistic associations between the visual tags and box weights through trial and error and to continuously adapt their predictions as contingencies changed throughout the task. Before starting the experiment, participants completed a short practice procedure using a visual example sequence of cue–outcome pairings to ensure that they understood the task structure and reversal procedure.

**Task 3: Effort-reward decision-making task**

Participants completed an effort–reward decision-making task adapted from established paradigms (*22–24*) to assess how exosuit assistance influenced effort-based choices and perceived fatigue (see Movie S3). The task consisted of 60 trials divided into four blocks of 15 trials, separated by brief rest breaks (30 s after trials 15 and 45, 60 s after trial 30). On each trial, participants chose between lifting a heavy, high-reward box and a light, low-reward box. The light box weighed 1.33 kg, whereas the weight of the heavy box was individually calibrated to 35% of each participant's MVF. Box weights remained constant across the task and across the assisted and unassisted sessions. The light box was associated with a fixed virtual reward of £1, whereas the heavy box was associated with a variable virtual reward ranging from £3 to £7. Heavy-box reward values were pseudorandomised such that each reward level was presented three times every 15 trials (12 times in total across the task). Participants were informed that a subset of trials would be randomly selected at the end of the experiment to determine a monetary bonus, although the specific trials were not disclosed in advance.

At the start of each trial, participants were informed of the reward associated with the heavy box and had 5 s to decide which box to lift. Participants lifted the selected box using the same technique as in the previous tasks and maintained the box at ~90° elbow flexion for 8 s. Trials were visually monitored by the experimenter, and participants were informed that trials would be invalidated if the required posture was not maintained. After returning the box to the table, participants verbally rated their perceived physical fatigue using a 0–100% scale, where 0 indicated feeling completely rested and 100 indicated the worst imaginable fatigue.

**Computational modelling**

<u>Reinforcement-learning models</u>

We fitted four reinforcement-learning models separately to each participant and assistance condition using custom scripts implemented in Python: a basic Rescorla-Wagner (RW) model, a cue-dependent RW model, an opponent RW model, and a Rescorla-Wagner–Pearce-Hall (RW-PH) hybrid model. Actual trial outcomes were coded as $O_t = 1$ for Heavy and $O_t = 0$ for Light. In all models, expected values ($V$) were initialised at 0.5, representing neutral initial expectations.

In the **basic RW model**, a single expected probability of the Heavy outcome was maintained irrespective of cue identity and updated after each observed outcome according to

$$V_{t+1} = V_t + \alpha(O_t - V_t),$$

where $V_t$is the expected probability of a Heavy outcome before observing the outcome on trial $t$, $O_t$ is the actual outcome, $O_t - V_t$is the prediction error, and $\alpha$ is the learning rate (0-1).

The **cue-dependent RW model** maintained separate expected values for cues A and B.

$$V_{c,t+1} = V_{c,t} + \alpha\left(O_t - V_{c,t}\right)$$

Here, $V_{c,t}$ is the expected probability of a Heavy outcome for the cue presented on trial $t$. Only the expected value associated with the presented cue was updated. A single learning-rate parameter $\alpha$ was shared across both cues.

The **opponent RW model** represented the instructed inverse relationship between the two cues using a single shared expected value. The recoded outcome $O_t^*$ was set to 1 for A-Heavy or B-Light trials, and 0 for A-Light or B-Heavy trials. The shared expected value was updated according to

$$V_{t+1} = V_t + \alpha(O_t^* - V_t)$$

The expected probability of Heavy was $V_t$ for cue A and $1 - V_t$ for cue B.

The **RW-PH hybrid model** additionally allowed the strength of learning to vary according to cue-specific associability:

$$V_{c,t+1} = V_{c,t} + \kappa a_{c,t}\left(O_t - V_{c,t}\right),$$
$$a_{c,t+1} = \eta \mid O_t - V_{c,t} \mid + (1 - \eta) a_{c,t}$$

Here, $a_{c,t}$ represents the current associability of the presented cue and was initialised at 1. The parameter $\kappa$ (0-1) scales the overall magnitude of learning, such that the effective learning strength on a given trial is $\kappa a_{c,t}$. The parameter $\eta$ (0-1) determines the extent to which associability is updated according to the magnitude of the most recent prediction error relative to its previous value. Thus, larger absolute prediction errors increase the influence of recent surprise on subsequent learning. As in the cue-dependent RW model, values and associabilities were cue-specific and only those corresponding to the presented cue were updated.

Response function and model fitting

In all four models, expected values were converted into response probabilities using a logistic response function:

$$P_t(\text{Heavy}) = \frac{1}{1 + \exp\left[-\left(b + \beta\left(V_{\text{cue},t} - 0.5\right)\right)\right]}$$

Here, $V_{\text{cue},t}$ is the model-derived expected probability of Heavy for the presented cue, $\beta$ is response sensitivity, determining how strongly learned expectations influence responses, and $b$ is response bias, allowing an overall preference for Heavy or Light responses independently of the learned expectation. This response function allows participants to make relatively categorical predictions rather than requiring their responses to probability-match their learned expectations.

The basic RW, cue-dependent RW and opponent RW models therefore each contained three free parameters $(\alpha, \beta, b)$, whereas the RW-PH model contained four $(\kappa, \eta, \beta, b)$.

Trial-wise expectancy responses were treated as binary observations, with model predictions specifying the probability of each response. Parameters were estimated by minimising the Bernoulli negative log-likelihood. Each model was fitted 100 times from randomly generated starting values to reduce sensitivity to parameter initialisation, and the solution yielding the lowest negative log-likelihood was retained. Models were compared using the Bayesian information criterion (BIC), which penalises the number of free parameters, with lower values indicating better fit.

Model selection and parameter recovery

Model comparison showed that the opponent RW model provided the lowest BIC in 55 of 64 participant-sessions (27 unassisted and 28 assisted), compared with eight for the RW-PH hybrid model, one for the cue-dependent RW model and none for the basic RW model.

Parameter recovery was assessed by generating simulated binary responses across a range of known learning rates ($\alpha$) and response sensitivities ($\beta$), with response bias ($b$) fixed at zero, and then refitting the opponent RW model. The learning-rate parameter showed good recovery ($r$ = .933, mean absolute error (MAE) = .105).

Following model comparison, learning-rate estimates from the opponent RW model were compared between assisted and unassisted conditions using a paired Wilcoxon signed-rank test.

Behavioural simulations

To assess whether the fitted opponent RW model could reproduce the observed behavioural effects, we simulated 1,000 datasets using each participant's fitted learning rate, response sensitivity and response bias, together with the cue-outcome sequence from each session. On each trial, the model-derived expectation was converted into the probability of predicting Heavy using the logistic response function. A Heavy or Light response was then sampled from this probability and scored as correct or incorrect using the same criteria as participants' actual responses.

Each simulated dataset was analysed using the same binomial mixed-effects model as the empirical behavioural data. We then compared the observed assistance-related differences in adaptation slopes with the distribution generated across the 1,000 simulations.

**Statistical analysis**

For all tasks, linear and generalised linear mixed-effects models were implemented using the *lme4* package in R. Participant was included as a random intercept. Categorical variables were coded using sum-to-zero contrasts and fixed effects were evaluated using Type III tests. Statistical significance was set at $p < .05$. Bayes factors were also calculated where appropriate, with $BF_{10}$ indicating evidence for the alternative hypothesis and $BF_{01}$ indicating evidence for the null hypothesis. Full model specifications in Wilkinson notation are provided in Supplementary Table S2.

## I. Weight perception task

Analyses were conducted separately for two box comparisons addressing conceptually distinct effects: the reference versus large box comparison assessed sensitivity to differences in actual mass, whereas the small versus large box comparison assessed size-related biases in weight perception between boxes of identical mass. Session order was included as a fixed effect to account for potential effects of the order in which the assistance conditions were experienced. MVF was included as a covariate to account for individual differences in participants' strength.

EMG amplitude was modelled as a function of box, assistance condition, trial number, and their interactions. MVF and its interaction with assistance were also included. An additional MVF × box interaction was included in the reference–large model, where the boxes differed in mass. An interaction between assistance condition and session order was also included. EMG amplitude was calculated across the entire trial for the main analyses. For the analysis of size-based anticipatory muscle activity in the size–weight illusion task, EMG amplitude was calculated over the lifting phase, from immediately before lift initiation until the box reached the target position.

Perceived-weight ratings were modelled as a function of box, assistance condition, trial, and their interactions. An MVF × box interaction and an assistance condition × session order interaction were additionally included.

## II. Probabilistic cue-outcome learning task

Prediction accuracy was modelled using binomial mixed-effects models, with trial-wise expectancy responses coded as correct or incorrect according to whether participants predicted the more probable box weight for the current cue–outcome contingency. Adaptation was quantified as the change in prediction accuracy across trials within each contingency block ("trials since reversal"). Because blocks differed in duration, primary analyses focused on the first 11 trials of each block, corresponding to the shortest block and allowing adaptation to be compared across equivalent learning and relearning windows. Prediction accuracy was modelled as a function of assistance condition, trials since reversal, contingency, and their interactions. Session order and an assistance condition × session order interaction were additionally included. Follow-up comparisons tested differences in adaptation slopes between assistance conditions separately for each contingency. To examine whether these effects differed across individual contingency blocks, a follow-up analysis modelled prediction accuracy as a function of assistance condition, trials since reversal, contingency block, and their interactions, again including session order and the assistance condition × session order interaction. Differences in adaptation slopes between assistance conditions were compared within each contingency block, with Holm correction for multiple comparisons.

III. Effort-reward decision-making task

EMG amplitude was modelled as a function of assistance condition, trial number, and their interaction. Load lifted was included to account for differences in the weights participants chose across trials and conditions. Session order and an assistance condition × session order interaction were additionally included.

Perceived fatigue was modelled as a function of assistance condition and trial number, and their interaction. Load lifted and pre-task fatigue (measured after the preceding task) were included as covariates, with the latter accounting for differences in perceived fatigue at the start of the effort–reward task. Session order and an assistance condition × session order interaction were additionally included. Trial-by-trial changes in perceived fatigue were additionally modelled using the same predictors, with EMG activity during the current lift and its interaction with assistance condition added.

Effort-based choices were analysed using a binomial mixed-effects model, with trial-wise choices coded as selecting either the low-effort or high-effort option. Choice was modelled as a function of assistance condition, trial number, session order, and their interactions. Reward magnitude was included as a covariate. Effort-based choices were additionally modelled as a function of assistance condition, trial number, cumulative EMG activity across preceding trials and perceived fatigue on the previous trial. Reward magnitude and session order were also included, together with the assistance condition × trial number and assistance condition × session order interactions.

Reward sensitivity was assessed using a binomial mixed-effects model, with choice modelled as a function of reward magnitude, assistance condition, trial number, and their interactions. Session order and an assistance condition × session order interaction were additionally included. To quantify evidence for the absence of a reward magnitude × assistance interaction, we calculated a BIC-approximated Bayes factor ($BF_{01}$) by comparing otherwise identical models with and without this interaction.

**Acknowledgements:** We are grateful to all our participants for their time and dedication. We thank Hristo Dimitrov and Enrica Tricomi for their technical insight during the design of the experiment and Gavin Buckingham for helpful advice on the design of the behavioural weight-estimation paradigm. **Funding:** The study was funded by UKRI's Frontier Research Guarantee (EP/X040372/1) and the Medical Research Council (MC_uu_00030/10). T.R.M. was partially funded by a Wellcome Trust Senior Research Fellowship (215575/Z/19/Z). N.W. was funded by an Israel Science Foundation Personal Research Grant (1603/22).
**Author Contributions**: M.M.-S and T.R.M. led the conceptualisation and experimental design of the study. N.W., R.M., F.M. and L.M. supported the experimental design. F.M. and L.M. designed and built the exosuit. M.M.-S, V.P., M.Z. and F.M. contributed to data collection. F.M. and J.R. provided technical support during data collection. M.M.-S and T.R.M. led the behavioural analysis with support from N.W. and R.M. M.M.-S, R.M. and T.R.M. led the computational analysis with support from M.S. F.M. led the EMG analysis with support from M.M.-S. M.M.-S and T.R.M. wrote the manuscript. M.M.-S prepared the figures. R.M., N.W., L.M. and F.M. provided feedback on the manuscript. T.R.M. and L.M. provided funding for the work. **Competing interests**: The authors declare that they have no competing interests. **Data and materials availability**: The data that support the findings of this study will be available from the Open Science Framework upon publication.

## SUPPLEMENTARY MATERIAL

### Supplementary results

### Opponent Rescorla–Wagner (RW) model reproduces the direction but not the magnitude of the adaptation effect

To assess whether the fitted opponent RW model could reproduce the observed behavioural effects, we simulated behaviour using each participant's fitted parameters and reanalysed the simulated data using the same mixed-effects model as the empirical data. For the A-light contingency, the mean simulated difference in adaptation slopes between unassisted and assisted conditions was 0.042 (95% simulation interval: −0.015 to 0.098), compared with the empirical difference of 0.126. In contrast, the empirical B-light difference (0.012) fell within the simulated distribution (mean = 0.027, 95% simulation interval: −0.035 to 0.090). Thus, although the opponent RW model reproduced the direction of the assistance effect for the A-light contingency, the simulated effect was smaller than the empirical effect.

## Supplementary figures

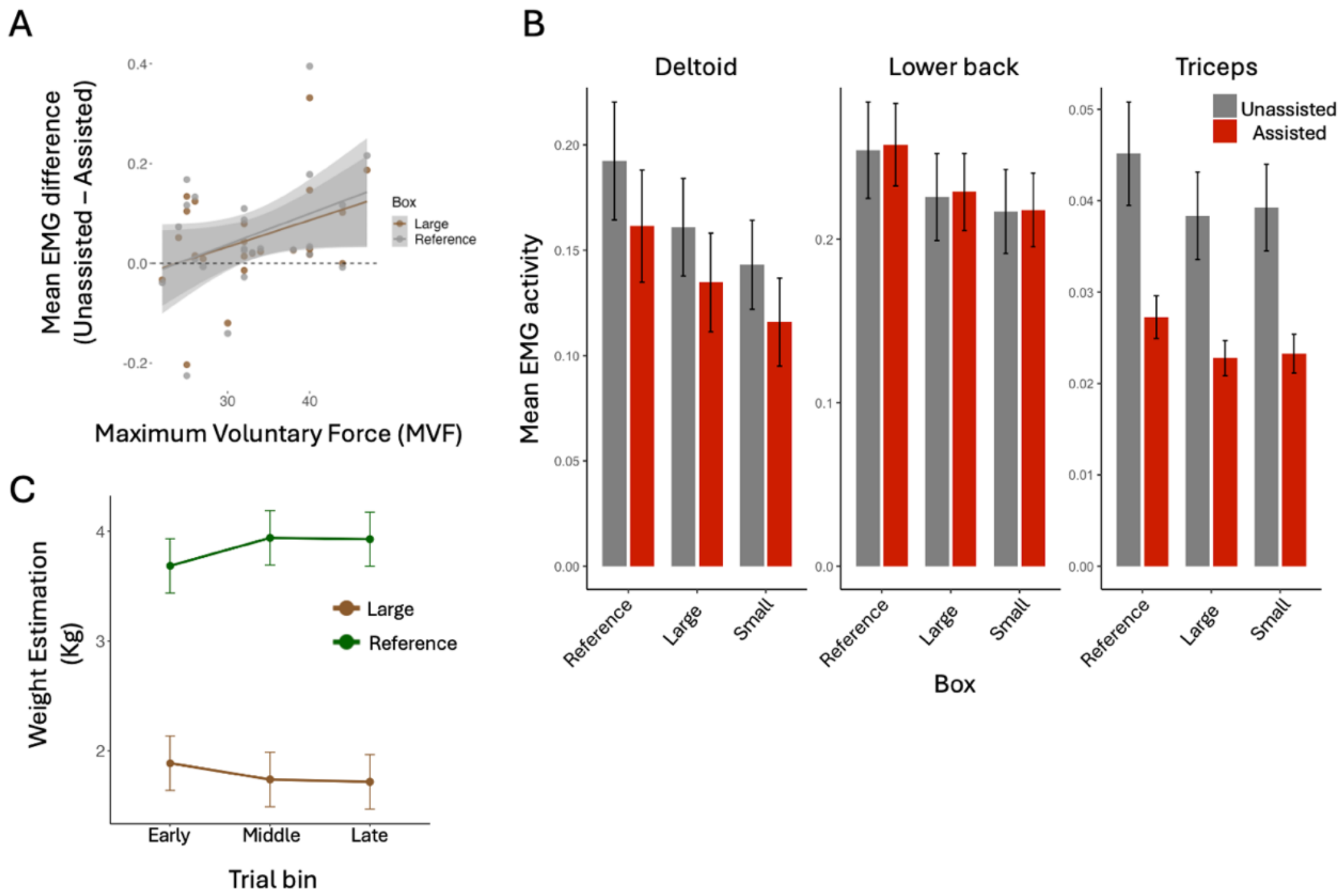

***Figure S1. Additional analyses of muscle activity and weight perception.***

*(**A**) Relationship between participant strength (maximum voluntary force; MVF) and the change in biceps EMG activity produced by exosuit assistance for the large and reference boxes. Positive values indicate lower EMG activity with assistance. Participants with higher MVF showed larger assistance-related reductions in EMG activity (MVF × assistance interaction: $F(1, 1355) = 185.62$, $p < .001$). Dots represent individual participants. Lines show linear fits with 95% confidence intervals. (**B**) EMG activity in the triceps, deltoid, and lumbar paraspinal muscles during the "hold" phase, examined for potential compensatory activity. Assistance did not increase activity in any muscle group: assisted − unassisted = −0.0193 for deltoid ($p < .001$), −0.00447 for lumbar paraspinal muscles ($p = .071$), and −0.0168 for triceps ($p < .001$). (**C**) Mean perceived weight estimates for the large (2.850 kg) and reference (3.860 kg) boxes across early (trials 1–5), middle (trials 6–10), and late (trials 11–15) trial bins, shown to illustrate the significant box × trial number interaction observed in the full mixed-effects model ($F(14, 1591) = 2.36$, $p = .003$). Weight estimates for the large box decreased slightly across repeated lifts, whereas estimates for the reference box increased, resulting in greater separation between the two boxes over time. Dots represent group means and error bars denote ± SEM.*

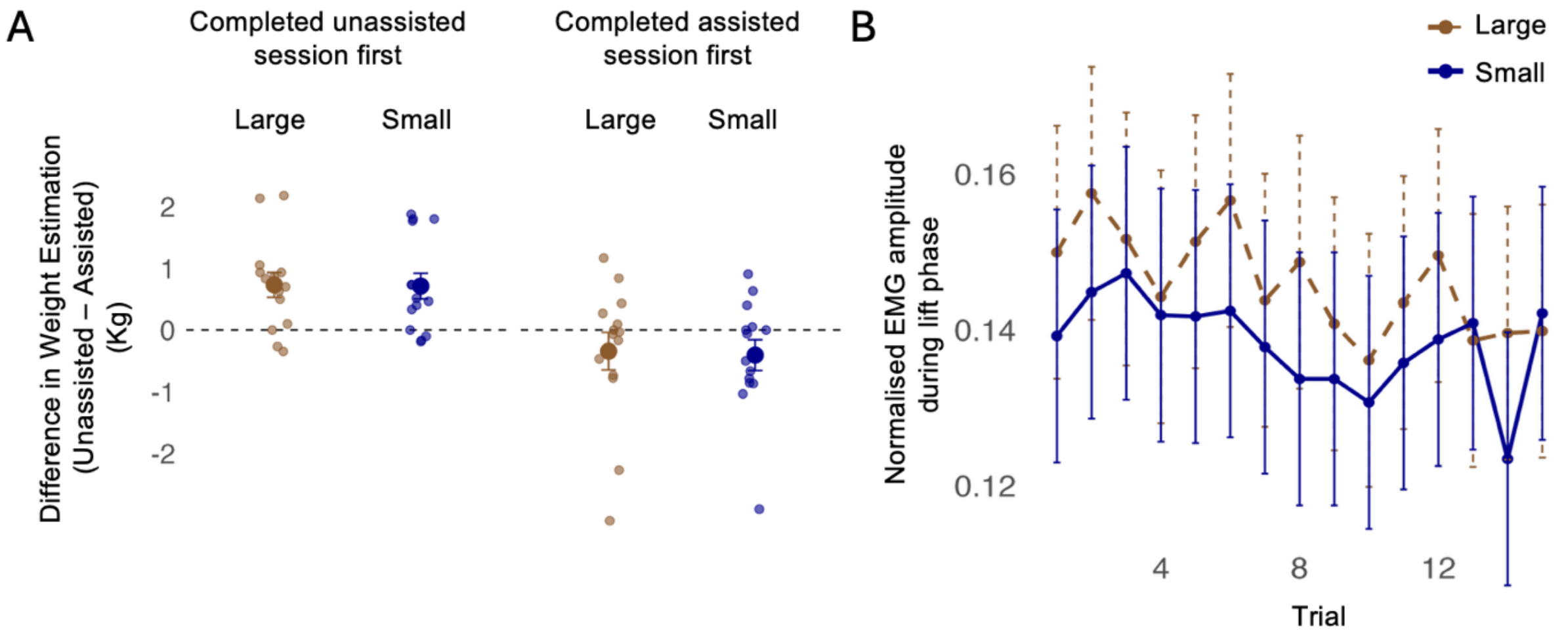

***Figure S2. Effects of exosuit assistance on weight perception and anticipatory muscle activity.***

*(**A**) Difference in perceived weight between unassisted and assisted sessions for the large and small boxes, shown separately according to whether participants completed the unassisted or assisted session first. Positive values indicate higher weight estimates during the unassisted session. The effect of assistance depended on session order, with both boxes perceived as heavier during the first experimental session regardless of whether it was assisted or unassisted (assistance × session-order interaction:* $F(1, 1592) = 198.95$, $p < .001$*). Large dots represent group means and error bars denote ± SEM; small dots represent individual participants. (**B**) Mean biceps EMG activity during the lift phase for the large and small boxes across repeated lifts. EMG activity was higher for the large than the small box (large − small = 0.008,* $F(1, 1414) = 5.41$, $p = .020$*), consistent with size-based anticipatory motor responses. This difference did not significantly change across repeated lifts (box × trial interaction:* $F(14, 1414) = 0.19$, $p = .999$*). Points represent group means and error bars denote ± SEM.*

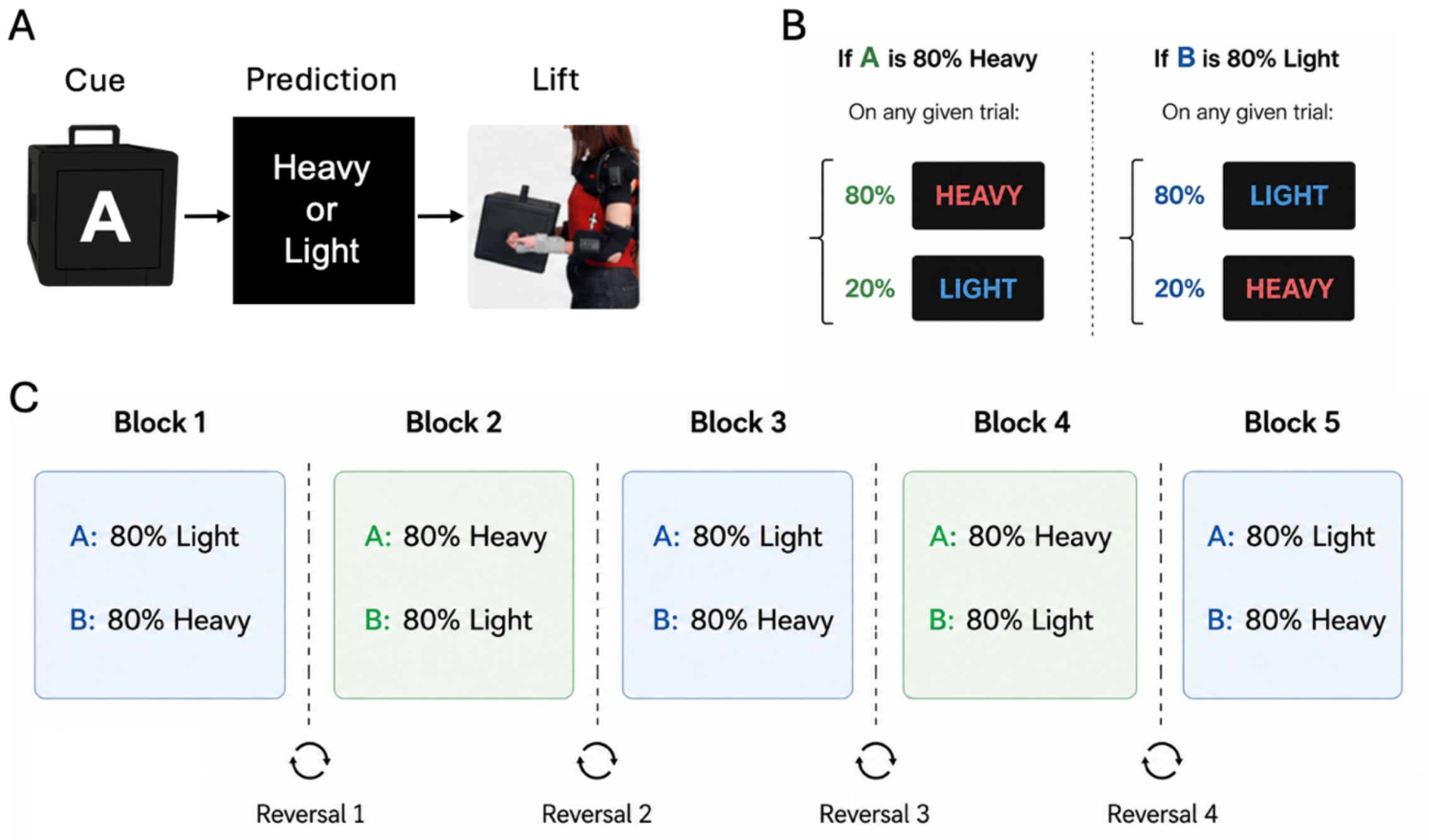

***Figure S3. Probabilistic cue–outcome learning task design and contingency reversals.*** *(**A**) Schematic of the probabilistic cue–outcome learning task. On each trial, participants viewed a cue (A or B), reported their prediction of whether the upcoming box would be heavy or light, and then lifted the box to experience its true weight. (**B**) Cue–outcome contingencies. At any given time, one cue predicted a heavy box on 80% of trials and a light box on 20% of trials, whereas the other cue predicted a light box on 80% of trials and a heavy box on 20% of trials. (**C**) Contingencies reversed between blocks, such that the cue–outcome associations alternated across five blocks. Participants were informed that the associations could change throughout the task but were not told when reversals would occur.*

## Supplementary tables

| Task | Acquired sample | Additional exclusions |
|---|---|---|
| Weight estimation task | N = 32 | N = 1 (poor engagement and inconsistent task responses) |
| Probabilistic cue-outcome learning task | N = 32 | N/A |
| Decision-making | N = 32 | N = 2 (pain during the task and fasting) |

**Table S1. Sample sizes and exclusions**

| Analysis | Model | Wilkinson notation |
|---|---|---|
| **Weight estimation task** | | |
| EMG – whole trial (reference–large) | LMM | EMG ~ Box × Assistance × Trial number + MVF × Assistance + MVF × Box + Assistance × Session order + (1 \| Participant) |
| EMG – lift-phase (small–large) | LMM | EMG ~ Box × Assistance × Trial number + MVF × Assistance + Assistance × Session order + (1 \| Participant) |
| Perceived weight – sensitivity to true mass (reference–large) | LMM | Weight ~ Box × Assistance × Trial number + Box × MVF + Assistance × Session order + (1 \| Participant) |
| Perceived weight – sensitivity to true mass controlling for EMG (reference–large) | LMM | Weight ~ Box × Assistance × Trial number + EMG + Box × MVF + Assistance × Session order + (1 \| Participant) |
| Perceived weight – size–weight illusion (small–large) | LMM | Weight ~ Box × Assistance × Trial number + Box × MVF + Assistance × Session order + (1 \| Participant) |
| **Probabilistic cue-outcome learning task** | | |
| Accuracy – contingency-level analysis | GLMM | Accuracy ~ Assistance × Trials since reversal × Contingency + Assistance × Session order + (1 \| Participant) |
| Accuracy – block-level analysis | GLMM | Accuracy ~ Assistance × Trials since reversal × Contingency block + Assistance × Session order + (1 \| Participant) |
| **Effort-reward decision-making task** | | |
| EMG | LMM | EMG ~ Assistance × Trial number + Load lifted + Assistance × Session order + (1 \| Participant) |
| Perceived fatigue | LMM | Fatigue ~ Pre-task fatigue + Assistance × Trial number + Load lifted + Assistance × Session order + (1 \| Participant) |
| Trial-to-trial change in perceived fatigue | LMM | Δ Fatigue ~ EMG × Assistance + Trial number + Load lifted + Pre-task fatigue + Assistance × Session order + (1 \| Participant) |
| Effort-related choices | GLMM | Choice ~ Assistance × Trial number × Session order + Reward + (1 \| Participant) |
| Choice accounting for accumulated muscle activity and perceived fatigue | GLMM | Choice ~ Assistance × Trial number × Session order + Cumulative EMG + Previous-trial fatigue + Reward + (1 \| Participant) |
| Reward sensitivity | GLMM | Choice ~ Reward × Assistance × Trial number + Assistance × Session order + (1 \| Participant) |

**Table S2. Statistical model specifications for each experimental task.**

| Variable | Coding |
|---|---|
| Assistance | Sum-to-zero: Unassisted = 1, Assisted = −1 |
| Session order | Sum-to-zero: Unassisted first = 1, Assisted first = −1 |
| MVF | Continuous, centred at the mean |
| Weight estimation | Continuous |
| Box | Sum-to-zero: Large = 1, Small = −1 / Large = 1, Reference = −1 |
| Trial number (weight estimation task) | Sum-to-zero: T1 = (1, 0,…, 0), T2 = (0, 1,…, 0), …, T14 = (0,…, 1), T15 = (−1, −1,…, −1) |
| Accuracy (probabilistic cue-outcome learning task) | Binary: Incorrect = 0, Correct = 1 |
| Trials since reversal | Continuous: 1–11, centred at the mean |
| Contingency | Sum-to-zero: A-light = 1, B-light = −1 |
| Contingency block | Sum-to-zero: B1 = (1, 0, 0, 0), B2 = (0, 1, 0, 0), B3 = (0, 0, 1, 0), B4 = (0, 0, 0, 1),<br>B5 = (−1, −1, −1, −1) |
| Fatigue | Continuous |
| Pre-task fatigue | Continuous |
| Δ Fatigue | Continuous: current-trial fatigue − previous-trial fatigue |
| Load lifted | Continuous |
| Choice | Binary: Low-effort box = 0, High-effort box = 1 |
| Reward (choice model) | Sum-to-zero: £3 = (1, 0, 0, 0), £4 = (0, 1, 0, 0), £5 = (0, 0, 1, 0), £6 = (0, 0, 0, 1), £7 = (−1, −1, −1, −1) |
| Reward (reward sensitivity) | Continuous, centred at the mean |
| Trial number (effort-reward decision-making task) | Continuous: 1–60, centred at the mean |
| EMG | Continuous |
| Cumulative EMG | Continuous: cumulative sum of EMG across preceding trials |
| Participant | Participant ID, random intercept |

**Table S3. Variables and coding used in the statistical analyses.**

**Supplementary movies**

**Movie S1. Perceptual weight estimation task.**

**Movie S2. Probabilistic cue-outcome learning task.**

**Movie S3. Effort-reward decision-making task.**

Supplementary Movies S1–S3 are available at https://osf.io/ft56y/files/osfstorage